\documentclass[conference, compsoc, twoside]{IEEEtran}

\usepackage{algorithm}
\usepackage[noend]{algpseudocode}
\usepackage{algorithmicx}
\usepackage{graphics}
\usepackage{adjustbox}
\usepackage{epsfig}
\usepackage{tablefootnote}
\usepackage{threeparttable}
\usepackage{booktabs}
\usepackage{amsmath}
\usepackage{algorithmicx}
\floatstyle{plain}
\newfloat{myalgo}{tbhp}{mya}
\usepackage{cite}

\begin{document}

 \title{Hand2Face: Automatic Synthesis and Recognition of Hand Over Face Occlusions}


\author{\IEEEauthorblockN{Behnaz Nojavanasghari}
\IEEEauthorblockA{Synthetic Reality Lab\\Department of Computer Science\\
University of Central Florida\\
Orlando, Florida\\
Email: behnaz@eecs.ucf.edu}
\and
\IEEEauthorblockN{Charles E. Hughes}
\IEEEauthorblockA{Synthetic Reality Lab\\Department of Computer Science\\University of Central Florida\\
Orlando, Florida\\
Email: ceh@cs.ucf.edu}
\\
\IEEEauthorblockN{Louis-Philippe Morency}
\IEEEauthorblockA{Language Technology Institute\\School of Computer Science\\Carnegie Mellon University\\Pittsburgh, PA\\
Email: morency@cs.cmu.edu}
\and
\IEEEauthorblockN{Tadas Baltru\v saitis}
\IEEEauthorblockA{Language Technology Institute\\School of Computer Science\\Carnegie Mellon University\\Pittsburgh, PA\\
Email: tbaltrus@cs.cmu.edu}
}
\maketitle
\begin{abstract}
A person's face discloses important information about their affective state. Although there has been extensive research on recognition of facial expressions, the performance of existing approaches is challenged by facial occlusions. Facial occlusions are often treated as noise and discarded in recognition of affective states. However, hand over face occlusions can provide additional information for recognition of some affective states such as curiosity, frustration and boredom. One of the reasons that this problem has not gained attention is the lack of naturalistic occluded faces that contain hand over face occlusions as well as other types of occlusions. Traditional approaches for obtaining affective data are time demanding and expensive, which limits researchers in affective computing to work on small datasets. This limitation affects the generalizability of models and deprives researchers from taking advantage of recent advances in deep learning that have shown great success in many fields but require large volumes of data. In this paper, we first introduce a novel framework for synthesizing naturalistic facial occlusions from an initial dataset of non-occluded faces and separate images of hands, reducing the costly process of data collection and annotation. We then propose a model for facial  occlusion type recognition to differentiate between hand over face occlusions and other types of occlusions such as scarves, hair, glasses and objects. Finally, we present a model to localize hand over face occlusions and identify the occluded regions of the face. 

\end{abstract}
 \begin{figure}[t!]   \includegraphics[scale = 0.7]{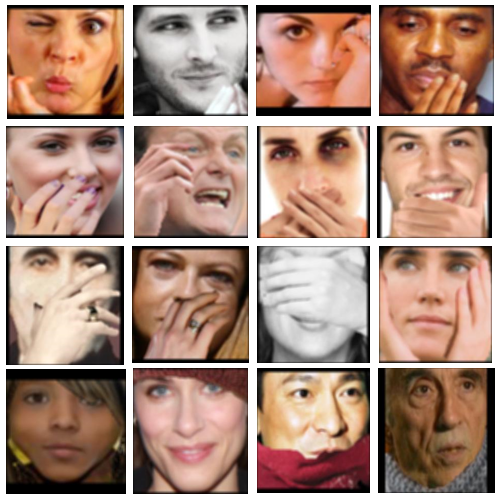}
      \centering
      \caption{Example synthetic images generated by our framework. As you can see, we are able to synthesize naturalistic hand over face occlusions as well as other common types of occlusion such as hair, hat and scarf occlusion. Hand over face occlusions contain one and both hand gestures that might happen in daily interactions.}
      \label{Figure 1}
   \end{figure}
\IEEEpeerreviewmaketitle

\section{Introduction}
In today's world people are spending a considerable amount of time in front of their computers. If our systems ignore some of the most important aspects of humans (e.g., social intelligence), valuable skills such as empathy, compassion and conflict resolution, which are the required skills to build and maintain any social relationship, can completely get lost in the rigid and long interactions with computers. Artificial social intelligence \cite{pantic2014social} is a step towards human-like human-computer interaction. Social intelligence is the ability to accomplish interpersonal tasks and respond to people appropriately \cite{bar2004emotional}.  Part of this ability depends on recognizing affective states of others in different situations, which can be through  visual, auditory, and tangible sensing\cite{thorndike1937evaluation}. Within the visual modality, facial expression analysis has been one of the most popular modalities to study affect in the affective computing community \cite{fasel2003automatic}. There has been a great amount of work on facial landmark localization but few of these include models with the ability to handle occlusion. Although there are some works on handling occlusion in landmark localization \cite{ghiasi2014occlusion ,ghiasi2015using,ghiasi2015occlusion}, the majority of these have focused on occlusions that are caused by hair bangs, sun glasses and other objects. While, these are some of the common facial occlusions, one of the most common types of facial occlusions are hand over face occlusions \cite{mahmoud2011interpreting}. These occlusions are particularly challenging because of the similar color and texture of the face and hands which can be misleading for current frameworks. Depending on the percentage of the facial pixels that are occluded, the region and type of occlusion, the face can be missed or facial expressions can be estimated incorrectly, which can lead to  false conclusions in affect recognition frameworks that heavily depend on facial expression analysis. 
 
Recognizing type of hand over face occlusions will not only benefit facial expression recognition frameworks, but also will be helpful in recognizing one's affective state \cite{d2017advanced,mahmoud2016automatic}. Hand over face occlusions and gestures are additional cues that play an important role in communicating one's affective state, while other types of occlusions such as hair, glass and hat are not meaningful in recognizing the affective state of a person.

One of the challenges for recognizing hand over face gestures is the lack of data and the time demanding and expensive process towards acquiring these datasets. To the best of our knowledge there is only one dataset of hand over face gestures \cite{mahmoud20113d}. However, this dataset does not cover all types of occlusions that can happen and are informative for emotion recognition. Also, in recognizing hand over face gestures, a small shift in location of the occlusion can result in a different interpretation of the gesture. Let us consider the hand over chin versus hand over nose gesture. While the first can indicate the thinking state of a person, the latter can be a cue for nervousness or deception\cite{reiman2007power,navarro2008every,caswell2003body}.  

In this paper, we (1) collect initial hand occluders as well as other types of occlusions that have been segmented from real images, (2) introduce a novel framework for synthesizing naturalistic facial occlusions and creating a massive corpus of naturalistic facial occlusions using the initial collected occluders set, (3) build models for facial occlusion type recognition to differentiate between hand over face occlusions and other types of occlusion, and finally we present a model to localize the hand over face occlusion region.  

\section{Related Work}

In this section we will review related work in facial occlusion recognition and gesture recognition areas as well as different approaches for synthesis and augmentation of data. 
\begin{figure}[t!]
    \includegraphics[scale = 0.45]{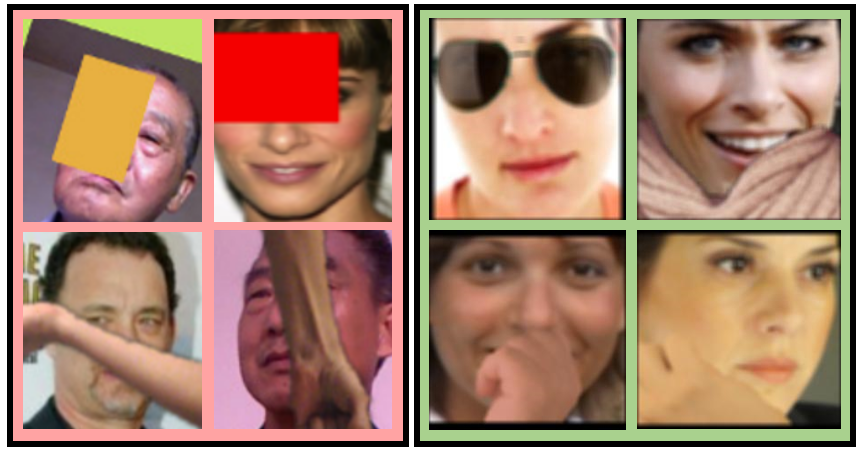}
      \centering
      \caption{Comparison between synthesized occluded faces using previous approaches and our approach. Pink box shows the synthesized examples using previous approaches\cite{saito2016real,ghiasi2014occlusion} and the green box shows our generated occluded faces.}
      \label{Figure 4}
   \end{figure}

Despite the importance of hand over face gestures\cite{mahmoud2011interpreting}, there has been limited research studying hand over face occlusions and hand gestures \cite{mahmoud2011interpreting,mahmoud2009towards,mahmoud2014automatic,mahmoud2016automatic}. 
In fact, we are only aware of two published research studies for automatic recognition of hand over face gestures \cite{mahmoud2014automatic,mahmoud2016automatic}.
Most work on facial occlusion has concentrated on building facial landmark detection and face recognition systems that are robust to it \cite{Ekenel2009,Yu2014,Saito2016,Wu2016}. 
Burgos-Artizzu et al. \cite{Burgos-Artizzu2013} explicitly detect which facial landmarks are occluded and include that information for better landmark location prediction.
Wu and Ji \cite{Wu2016} iteratively predict landmark occlusions and landmark locations using an occlusion pattern as a constraint.
Saito et al. estimate a facial occlusion mask by using a convolutional neural network trained on synthetic data \cite{Saito2016}.
However, all of the above mentioned work treat all types of occlusions in the same manner and do not differentiate between occlusions by hands and other objects.
In contrast, our work is able to distinguish between different types of facial occlusions.

Large corpora make a huge difference when training and building models in every research area. However, the process for collecting affective data in naturalistic settings and annotations is time demanding and expensive. In order to increase the number of data points, which results in more robust and generalizable models, some researchers augment real-world data with synthetic data. Synthetic data are specifically popular in creating occluded faces, where random sized rectangles are placed on different regions of the face \cite{ghiasi2014occlusion,ghiasi2015using,ghiasi2015occlusion}. Although this procedure helps with augmenting data and is useful in training more accurate models for landmark localization, face segmentation and facial occlusion localization, synthetic occluded data do not include hand over face gestures. Also, as the goal of these studies is not affect recognition, naturalness of the synthetic data was not one of the factors in generation of the synthetic data. However, when it comes to inferring affective states from faces that include occlusions, we need naturalistic data that contain a variety of hand over face gestures. Figure 2 shows a comparison between generated occlusions in previous works and our approach.  These occlusions are particularly challenging to recognize as face and hand have similar color and texture and can challenge the performance of previous previous frameworks\cite{baltru2016openface,burgos2013robust} in detecting faces and recognizing facial expressions. 

Although previous studies have taken initial steps towards addressing facial occlusions and recognizing hand over face gestures \cite{mahmoud2009towards}, they have not differentiated between the facial occlusion types. Differentiating the type of the occlusion, whether it is caused by hand or other objects, can also be informative for affective recognition. 

\section{Synthesis Pipeline}
Our synthesis pipeline contains several steps towards generating the synthetic image. Figure 4 shows the overview of our pipeline. We first start with segmenting the hand (or other) occlusions from a real occluded face to get the initial occluder. After the segmented occluders are obtained, we retrieve a face image from the database on which we want to generate the occlusion. Then we proceed to do color correction, quality matching and re-scaling of the occluder according to a the retrieved face. Finally, we generate the occlusion at the desired region of the face and get the corresponding occluder. Algorithm 1 briefly describes our synthesis process.
 \begin{figure}[t!]
   \includegraphics[width = 0.4 \textwidth , height= 0.6\textheight]{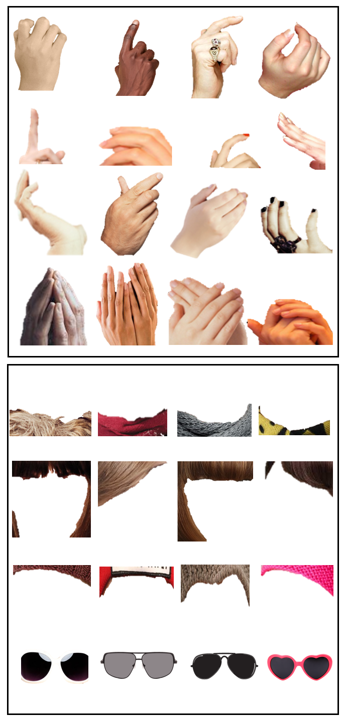}
      \centering
      \caption{Example of initial occluder in our dataset. The first box corresponds to hand occluders that can cause minor occlusions as well as hands that can cause heavy to full occlusions. The hand occluders include one hand and two hand gestures. The second box corresponds to other categories of occlusions such as scarf, hair bangs, hat and glasses.}
      \label{Figure 2}
   \end{figure}

\begin{algorithm}[t!]
\caption{Synthesis of naturalistic facial occlusions}\label{euclid}
\begin{algorithmic}[1]
\For{\textit{ \textless all faces in database \textgreater}}
\State  $\textit{occluder} \gets pose\_match(face)$ 
\State  $occluder \gets {scale(occluder)}$
\State $region \gets {region\_encoding(face,
 landmarks)}$
\State $occluder \gets {match\_color\_illumination(face,occluder)}$
\State  $occluder \gets {match\_quality(face,occluder)}$
\State $syn\_img \gets {composit (face,occluder,region)}$
\EndFor 
\\
\Return $syn\_img$
\end{algorithmic}
\end{algorithm}
\subsection{Initial Occluder Acquisition}
In order to be able to synthesize facial occlusions we start from an (1) initial set of non-occluded faces, and (2) initial set of potential occluders that contain hand occluder as well as other types of occluders such as hair, scarves, hats, etc. that are crawled from the web. To achieve the first goal we have selected non-occluded faces from the LFPW dataset \cite{belhumeur2013localizing}. Our choice of dataset was based on the fact that LFPW contains faces in the wild that have different facial expressions, various poses and were captured under varying lighting conditions, which can result in a more diverse and near to real world synthetic dataset. To obtain the second goal, an initial set of occlusions, we have downloaded images of occluded faces from the web. These occlusions include hand over face gestures, glasses, hair bangs, hats and scarves. As the goal of creating the synthetic data is for the purpose of affect recognition studies, we have chosen specific words that are motivated by previous studies on affect recognition and hand over face gesture studies \cite{mahmoud2011interpreting} \cite{monkaresi2016automated,nojavanasghari2016emoreact,nojavanasgharifuture,nojavanasghari2017exceptionally}.
Examples of key words we have used are as follows: Hand over chin, hand on mouth, frustrated person, bored person, worried person, etc.
We have also augmented our hand occluder using 50 additional occluders from the COFW dataset (training set) \cite{burgos2013robust}, which was originally collected to study problems related to facial occlusions. 
\subsection{Occlusion Segmentation}
We have done a manual cleaning of the downloaded data for quality assessment. Since getting a clean occluder is necessary for the success of subsequent stages of the synthesis algorithm, we have used interactive grabcut segmentation \cite{rotherinteractive} to segment the hand occluders from the original images. This framework allows the user to guide the process in a couple of loops during the segmentation , which is crucial for our challenging segmentation problems such as segmenting hand from face. 

 \begin{figure*}[!th]
    \includegraphics[width =1 \textwidth]{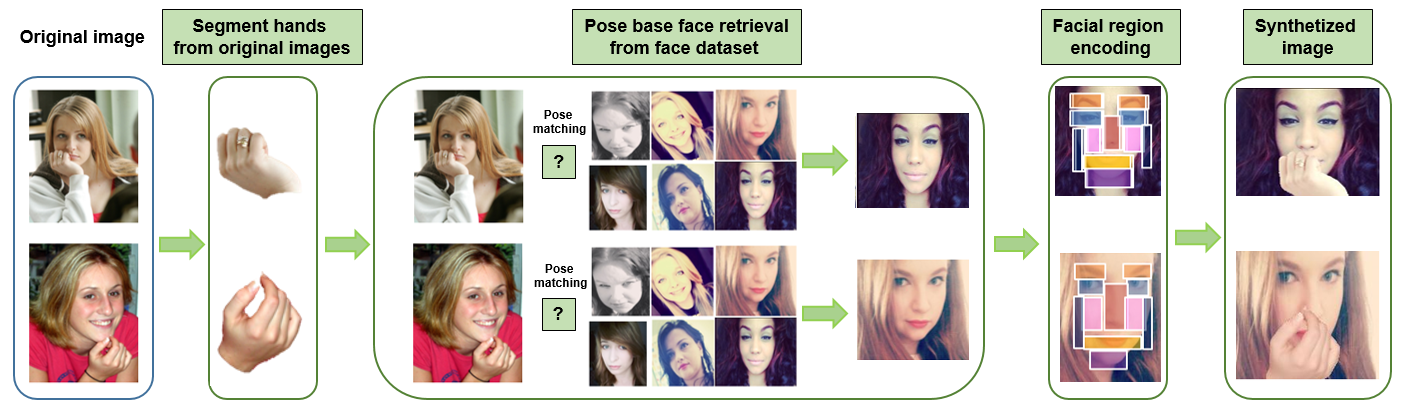}
      \centering
      \caption{Our synthesis pipeline. }
      \label{Figure 3}
   \end{figure*}

\subsection{Matching Faces with Occlusions}
Our synthesis pipeline is made of the following four steps (summarized in algorithm 1): For each non-occluded face from the database, we retrieve a suitable occluder from an initial occluders set, re-scale the occluder based on the face's size, match the the color, illumination and quality  of the face and the occluder, and finally put the occlusion on a selected area of the face. 
 
 \begin{figure}[b!]
    \includegraphics[scale = 0.5]{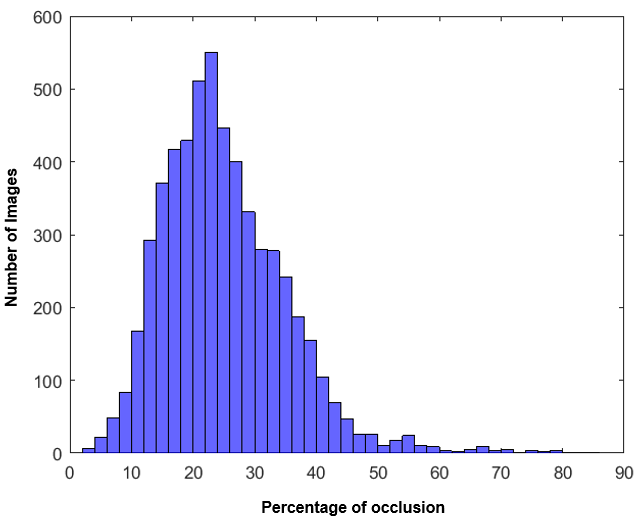}
      \centering
      \caption{Distribution of occlusion in our synthetic dataset.}
      \label{Figure 5}
   \end{figure}


\subsubsection{Pose and orientation}
Some hand gestures can only happen with certain orientations of the face ( See Figure 1).  Therefore, before synthesis of hand over face gestures we match the orientation of the original face from which we have segmented the occluder, with the target face on which we want to generate the occlusion based on yaw, roll and pitch of the face.  We retrieve top matching faces from the database based on pose and randomly pick a target face image from the top 10 matched images.

\subsubsection{Scale}
One of the most important factors of naturalistic synthesis of facial occlusion is scale. In our synthesis, we re-scale the occluder based on the face size to generate naturalistic gestures.

\subsubsection{Color correction}
The hand and face should match in terms of color and illumination to be perceived as natural. The goal of this step is to make the occluding hand take on the look and feel of the facial image. Following the previous work in color transferring between images \cite{shirley2001color}, we compute the mean and standard deviation across each channel of the face image and impose the same mean and standard deviation on the occluder image. This step will transfer the appearance of the face to hand occluder and will generate naturalistic synthetic results. 

\subsubsection{Quality of the image}
 One of the factors for naturalness of the image is matching the quality of the images of target face and hand occluders; ideally they should match precisely. To make sure that this criterion is satisfied, we check the sharpness of the face and hand occluder using previously defined metrics \cite{crete2007blur}. Then we constrain the face and occluder to have the same sharpness.

\subsection{Region Encoding of the Face}
In order to synthesize the occluded faces, we need to know the area of the face on which we will place the occlusion. We have decided to segment the face into 11 regions, representing the major face parts (i.e. left and right margins, left and right cheeks, left and right eyes, left and right eyebrows, nose, mouth and chin). These areas can be coded based on the facial landmarks that are the input to our framework. As we have selected faces from the LFPW dataset\cite{belhumeur2013localizing}, the ground truth landmarks are provided along with the faces. Note that, in generating the occluded faces, a Gaussian noise was added to these areas to achieve more robustness

\subsection {Compositing Naturalistic Facial Occlusions}
After obtaining the occluders and matching them in terms of scale, color and orientation, our framework synthesizes occluded face images, placing the occluders on different areas of the face.

\subsection{Statistics of Hand2Face Dataset}
\textbf{Faces:} We started by selecting 431 non-occluded faces from the LFPW dataset\cite{belhumeur2013localizing} as our target faces on which we generate the occlusions. 

\textbf{Occluders:} In total we have 198 hand occluders (148 crawled occluders from the web and 50 augmented from the COFW dataset). For other types of occlusions, we have 328 occluders (15 glasses, 50 hair bangs, 20 hats, 10 scarves and additional occluders segmented from using the COFW dataset as a training set that includes our categories as well as other objects). Despite other datasets that have been created for studying facial occlusions \cite{burgos2013robust}, we have not put a restriction on percentage of facial occlusions, pixel ratio of occluded areas of the face to the entire face area, to simulate the real world examples. Figure 5 shows the facial occlusion percentage distribution in our synthetic data. We have generated 9912 occluded images and on average there is 25\% occlusion on generated faces and the minimum and maximum percentage of occlusions are 3\% and 85\%.

\section{Facial Occlusion Type Recognition and Localization}
Facial occlusion type can encode valuable information about a person'€™s emotional state. Differentiating types of occlusions can also be informative for affect recognition; however, there is no known technique to differentiate between facial occlusions. Given our large corpus of data with various combinations of hand over face and other types of occlusions across different individuals and conditions, our goal is to learn an effective representation and build computational models to classify the type of facial occlusions.  
\subsection{Experimental Methodology}

\textbf{Feature Representation:} Considering the popularity and success of convolutional neural networks \cite{krizhevsky2012imagenet} in various computer vision problems such as image classification, we decided to adopt these networks to extract meaningful representations for occluded faces. The last layer of these networks is a good representation for our task as it contains learned textures, colors, shapes and even parts of various faces as well as other objects that could potentially be  occluders in our problem.  We used a pre-trained very deep convolutional network (VGG net) \cite{simonyan2014very} that was trained on the ImageNet large scale image dataset \cite{fei2010imagenet}.  We kept all the layers up to the softmax layer and extracted features from the last layer as a way of representing each occluded face. Thus, our feature representation is a vector of dimension $1\times1000$ for each occluded face.
\newline\textbf{Implementation Details:} We started with 9912 occluded faces, from which 5172 images were hand over face occlusions, and 4740 images were other types of occlusions (e.g. hair, scarf, etc.), which are generated by our synthesizer.  We used 40\% of the data for training, 30\% for validation and 30\% for testing.  We used majority voting, naive Bayes, radial basis function kernel SVM  \cite{chang2011libsvm}, binary logistic regression \cite{bishop2007pattern} and a deep neural network \cite{nojavanasghari2016deep} as our baseline models. We used the validation set for selecting the hyper parameters of our models. For choosing parameters of SVM we did a grid search for $\gamma$ and $C$ on $ \gamma= \{2^-5,...,2^5\}$  and $C=\{10^-5,...,10^5\}$ sets. For choosing the values of the deep neural architecture we did a grid search on the number of units, number of layers, dropout rate and number of epochs over $units = \{16,32,64,128,256,512\}$, number of layers = $\{2,3,4,5,6,7,8\}$, dropout = \{0.1,0.3,0.5,0.7\}, and $epochs = \{30,50,70,100\}$ values. We chose Adam as our optimizer in DNN \cite{kingma2014adam}. Cross validation for choosing these values has been shown to be more effective than fixing them \cite{nojavanasghari2016deep}.
\begin{table*}[t!]
\caption{Facial Occlusion type Recognition. Our proposed DNN performs better than baseline models.}
\label{table_example}
\begin{center}
\begin{tabular}{|l||cccc|}
\hline \textbf{
Method} &\textbf{Accuracy} &\textbf{precision} &\textbf{ Recall}&\textbf{ F1} \\*
\hline
Deep Neural Networks (CNN features) &\textbf {\textit{85.36}} &\textbf{ \textit {84.63}} &\textbf{ \textit {89.11}} &\textbf{ \textit {86.81}}\\
Binary Logistic Regression (CNN features) & 82.27& 82.25 & 82.00 & 82.12\\
Support Vector Machine (CNN features) & 82.41&82.34 & 82.20& 82.26\\
Naive Bayes (CNN features) &70.43& 68.78& 73.20& 70.92 \\
Majority Voting & 54.05 & 54.05 & 100&70.12\\
\hline
\end{tabular}
\end{center}
\end{table*}
 \begin{figure*}[th!]
    \includegraphics[scale=0.5]{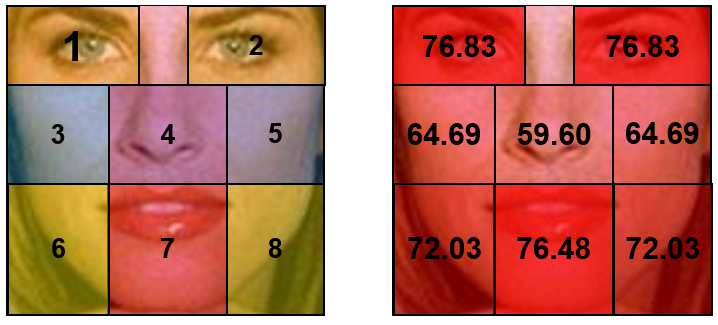}
      \centering
      \caption{Hand over face occlusion localization results. On the left we show how we divide the face into 8 regions and on the right you can see the accuracy of occlusion detection in each region. }
      \label{Figure 6}
   \end{figure*}
\subsection{Results and Discussions}
In our initial experiments for occlusion type classification, hand over face occlusion vs. other types of occlusion, we used several baseline models that are described in this section. Table 1 shows the results of our baseline models for facial occlusion type recognition. Our input features to the models are CNN features, which, given an input image, represent information regarding color, text, shape and parts of the faces and their occlusions. Ideally the classifier should be able to draw conclusions about facial occlusion type, using a combination of factors both from the face and its occluder. In our experiments deep neural networks achieve the best performance among our baseline models. This model uses a non-linear function at every layer and multiple layers add levels of abstraction that cannot be contained within a single layer. While the first layer is a non-linear function of linear combinations of inputs, the higher layers provide more complex and richer knowledge than the initial ones, which results in better performance.

In our experiment for hand over face occlusion localization, we first divided the face into 8 different regions (See Figure 6). We extracted CNN features from each region and used the deep neural network classifier to classify each region as an occluded/non-occluded area. Our results suggest that recognizing hand over face occlusions when they occur on eye and mouth regions are more successful compared to other areas. This is consistent with the statement that previous works have made about the more discriminatory power of these regions \cite{ekenel2009facial}. When these areas are occluded, it is easier for the classifier to notice their absence. On the other hand, for areas on the borders of the face and nose, it is more challenging to recognize that they are occluded, as hand and face have similar color and texture and hand occlusions can be considered as a part of these areas. 
\section{Conclusions}
Facial occlusions have always posed different challenges in   areas of affect recognition, face recognition and facial expression analysis and landmark localization. These occlusions are usually treated as noise, while they can provide helpful information towards the goal of affect recognition. Also, the performance of current facial recognition frameworks are often challenged due to facial occlusions; specifically if the occlusion is caused by hands. 
In this paper, we first introduced a novel framework for synthesizing naturalistic facial occlusions that include hand over face occlusions as well as other occlusion types. We introduced a new problem as facial occlusion type recognition that can be used as an additional cue towards recognizing one's affective state. We built several baseline models for recognition of facial occlusion type, where deep neural networks outperformed other baselines, confirming the effectiveness of these architectures. Finally, we used our deep neural network architecture to localize the occluded regions in hand over face occlusions. Our results showed that hand over face occlusions are easier to recognize when they occur on facial parts that have more discriminative power such as eyes, eyebrows and the mouth area, while they are more challenging when they happens on borders of the face and nose area. 


\section*{Acknowledgment}
This material is based upon work partially supported by The Heinz Endowments and the Bill \& Melinda Gates Foundation(OPP1053202). Any opinions, findings, and conclusions or recommendations expressed in this material are those of the author(s) and do not necessarily reflect the views of The Heinz Endowments or the Bill \& Melinda Gates Foundation, and no official endorsement should be inferred.


\bibliographystyle{IEEEtran}
\bibliography{acii.bib}

\end{document}